\documentclass{article}

\usepackage{microtype}
\usepackage{graphicx}
\usepackage{subcaption}
\usepackage{multirow}
\usepackage{booktabs} %

\usepackage{hyperref}

\usepackage[accepted]{icml2026}

\usepackage{amsmath}
\usepackage{amssymb}
\usepackage{mathtools}
\usepackage{amsthm}

\usepackage[capitalize,noabbrev]{cleveref}

\theoremstyle{plain}

\theoremstyle{definition}

\theoremstyle{remark}

\usepackage[disable,textsize=tiny]{todonotes}

\icmltitlerunning{Linear-LLM-SCM: Benchmarking LLMs for Coefficient Elicitation in Linear-Gaussian Causal Models}

\begin{document}

\twocolumn[
  \icmltitle{ Linear-LLM-SCM: Benchmarking LLMs for Coefficient Elicitation in Linear-Gaussian Causal Models }

  \icmlsetsymbol{equal}{*}

  \begin{icmlauthorlist}
    \icmlauthor{Kanta Yamaoka}{dfki,rptu}
    \icmlauthor{Sumantrak Mukherjee}{dfki}
    \icmlauthor{Thomas Gärtner}{hpi}
    \icmlauthor{David Selby}{dfki}
    \icmlauthor{Stefan Konigorski}{hpi,mountsinai}
    \icmlauthor{Eyke Hüllermeier}{dfki,lmu,mcml}
    \icmlauthor{Viktor Bengs}{dfki}
    \icmlauthor{Sebastian Vollmer}{dfki,rptu}
  \end{icmlauthorlist}

  \icmlaffiliation{dfki}{Data Science and its Applications, German Research Centre for Artificial Intelligence (DFKI), Germany}
  \icmlaffiliation{rptu}{Dept. of Computer Science, University of Kaiserslautern--Landau (RPTU), Germany}
  \icmlaffiliation{hpi}{Digital Health - Machine Learning Research Group, Hasso Plattner Institute for Digital Engineering, Germany}
  \icmlaffiliation{lmu}{Institute of Informatics, University of Munich (LMU), Germany}
  \icmlaffiliation{mountsinai}{Hasso Plattner Institute for Digital Health at Mount Sinai, Icahn School of Medicine at Mount Sinai, USA}
  \icmlaffiliation{mcml}{Munich Center for Machine Learning (MCML), Germany}

  \icmlcorrespondingauthor{Kanta Yamaoka}{kanta.yamaoka@dfki.de}

  \icmlkeywords{Machine Learning, ICML}

  \vskip 0.3in
]

\printAffiliationsAndNotice{}  %

\begin{abstract}
Large language models (LLMs) have shown potential in identifying qualitative causal relations, but their ability to perform quantitative causal reasoning---estimating effect sizes that parametrize functional relationships---remains underexplored in continuous domains. We introduce Linear-LLM-SCM, a plug-and-play framework for evaluating LLMs on Linear Gaussian structural causal model parametrization when a directed acyclic graph (DAG) is given.
The framework decomposes a DAG into local parent-child sets and prompts an LLM to produce a regression-style structural equation per node, which is aggregated and compared against available ground-truth parameters.
Our experiments with seven real-world DAGs effect ground truth  illustrate limitations of LLMs as quantitative causal parameterizers. Across most models, we observe variability in coefficient estimates and sensitivity to structural perturbations. We open-sourced the framework to further encourage the community to work on studies toward the use of LLM for causal effect elicitation in safety-critical domain, e.g., healthcare.  

\end{abstract}

\section{Introduction}

Robust intelligence requires an agent to have an internal ``world model''~\citep{ha2018worldmodels}, that is, an internal causal mechanism to infer causal structures and their effect relationships~\citep{Pearl2019causalinferenceandml}. Recent work~\citep{richens2024robustagentslearncausal} suggests that \emph{any agent capable of solving complex decision tasks must effectively learn a causal model of its data-generating process.} Given the advancement of LLMs, they exhibit the ability to 
encode a corpus of human knowledge,
e.g., clinical knowledge~\citep{Singhal2023}. One may speculate whether LLMs encode literature with causal information and constructing causal models. Per \emph{LLMs' Causal Hierarchy}, classified by \citet{zhang2023understandingcausalitylargelanguage}, LLMs have shown promise in Type 1 tasks (identifying causal relationships using domain knowledge), but they struggle with Type 2 (discovering new knowledge from data) and Type 3 (quantitative estimation of consequences) tasks.
Empirically and theoretically, Type 3 aspects in continuous domains remain under-explored. To fill this gap, we propose
\emph{Linear-LLM-SCM}, a 
framework 
evaluating quantitative causal capabilities of LLMs. It decomposes DAGs into local parent-child structures, assigning LLMs to elicit regression coefficients for linear Gaussian SCMs.
In our open-sourced software, researchers can use their own DAGs and LLMs plug-and-play to evaluate in their domains. %

In this contribution, we investigate the following \textbf{Research Questions (RQs)}: \textbf{RQ1:} Given a pre-specified DAG, can LLMs elicit plausible regression coefficients compared to real-world ground truths?
\textbf{RQ2:} How robust is this parameterization when facing adversarial conditions, such as DAG misspecification (spurious edges) or changes in variable units?
\textbf{RQ3:} What are the common failure modes encountered when using LLMs for parameterization?

Structural perturbation in RQ2 accounts for observational or modeling limitations. Real-world modeling, including the specification of a directed acyclic graph (DAG) as ground truth, contains uncertainty~\citep{padh25assumeddagiswrong, vowels2023dagsurvey}.  If possible, subsequent causal effect estimation should be robust against them. This also mitigates data leakage in the LLM training corpus~\citep{yang2023criticalrevcausalinfbenchllm}.

\textbf{Contributions.} We introduce a framework to evaluate LLMs' ability to estimate linear SCM coefficients. %
We open-source an evaluation pipeline that supports plug-and-play DAGs, variable metadata, and 
LLMs, reporting coefficient-distance and ordering metrics against ground truth.
~\footnote{\href{https://github.com/datasciapps/parameterize-dag-with-llm}{https://github.com/datasciapps/parameterize-dag-with-llm}}

\section{Background and Related Work}

Structural causal models (SCMs) formalize causal systems through a set of structural equations (e.g., in form of a DAG) and distributional assumptions.
Recent work explored using LLMs 
for causal reasoning, focusing on causal discovery and qualitative inference \citep{long2024llmsbuildcausal,long2023causaldiscoverylanguagemodels,kıcıman2024causalreasoninglargelanguage}. Recent studies raise concerns about LLMs’ causal capabilities, arguing that they may succeed at identifying causal relationships using prior knowledge, while struggle to discover causal structure from data and quantitatively estimate causal %
effects~\citep{zhang2023understandingcausalitylargelanguage,zecevic2023causalparrotslargelanguage,yang2023criticalrevcausalinfbenchllm,jin2024cladderassessingcausalreasoning}.

Existing work on LLM-based causal effect estimation focuses on discrete domains or requires numerical observational data and specialized architectures \citep{chen2023distilincounterfactualswithllms,feder2024dataaugmentationsimprovedlarge,zhang2024causalfoundationmodelduality}. Closest to our setting,  \citet{bynum2025languagemodelscausaleffect} combined LLMs with pre-specified causal graphs but estimate effects via sampling-based conditional distributions rather than directly eliciting structural parameters. In contrast, we study whether LLMs can directly estimate \emph{continuous} linear causal effect parameters for pre-specified SCMs using only causal structure and variable semantics, without observational data. Appendix~\ref{app:relatedwork} provides a detailed literature review.

\section{Linear-LLM-SCM Framework}

We provide pre-specified DAG structures decomposing real-world phenomena into parent-child interactions. The system iterates through the DAG, calling LLMs via prompt templates to elicit functional mappings for each parent-child set, then aggregating results to obtain full effect parameters. This assumes that LLMs have learned quantitative or qualitative information from their large training corpus. For quantitative aspects, LLMs may have encountered scientific literature with effects formalized as SCMs, 
or linear regressions.
For qualitative aspects, LLMs encountered natural language causal statements (e.g., ``Coffee consumption positively regulates alertness after 1 hour'') in their corpus.

\subsection{Overview of the benchmarking framework}

Our benchmarking framework requires a DAG structure $\mathcal{G}$ consisting of nodes $\mathcal{V}$ and directed edges $\mathcal{E}$, variable descriptions $\mathcal{D}$ including a short textual description and their unit (e.g., ``GC'': ``Glucose ($\mu$M)''), and variable constraints $\mathcal{R}$ (lower and upper bounds). The parameterization task assumes linearity in structural equation, Gaussian noise of target variables while beta coefficients do not have noise assumption. The variable constraints were sourced from original literature or author judgment, as shown in Appendix~\ref{app:dag_selection} Table~\ref{tab:dag_table}. The DAG structure is given by a yaml file containing that information and effect ground truths. An example including the required input can be found in Appendix~\ref{app:benchmark_impl_details} in Figure~\ref{fig:yaml_dag_configuration}. Given these inputs, Linear-LLM-SCM traverses nodes in topological order and decomposes the graph into local parent-child elicitation tasks. For each target node, it prompts the LLM with domain context, variable descriptions, units, ranges, and output-format instructions, then parses the returned linear equation to extract the intercept and parent coefficients. %

\textbf{Response Parsing:} We obtain the functional mapping of direct parents and target variables as textual structural equations from LLMs. While LLMs provide plausibility descriptions for debugging, we utilize only the parameterized equation provided as a string.
From this property, the system extracts beta coefficients from the LLM structural equation format, which should be compliant with the demonstration in the prompt template as in Appendix~\ref{app:framework_details} Figure~\ref{fig:prompt_template}. %

\subsection{Prompt for Node Level Function Parametrization} \label{sub:prompt_template}

Our framework traverses a DAG.
For each node, the LLM receives descriptions of its direct parents. %
While one could pass the entire graph via text representations (e.g., DAGitty by ~\cite{dagtty2017} or Mermaid by~\cite{mermaid2014}) or feed entire DAGs into visual language models, we chose the parent-child template instead. This focus on local structures aligns with recent work~\citep{bynum2025languagemodelscausaleffect, nafar2025extractingprobabilisticknowledgelarge}.

Our prompt includes the names and short descriptions of the target variable and its direct parents, alongside formatting instructions. On a high level, the prompt consists of three parts: First, domain expert persona (e.g., expert in consumer behaviour), a summary of the phenomenon of interest (e.g., `` Cachexia is a complicated metabolic syndrome related to...''), and variable units are introduced to the LLM. Next, the parameterization task, linear equation template, and variable ranges appear. Finally, the LLM is informed of the output format, which starts with the thought process (\emph{reasoning tokens}) and ends with the parameterization result as a string.  Appendix~\ref{app:framework_details} Algorithm~\ref{alg:benchmarking} includes details of this process as line 4, and Figure~\ref{fig:prompt_template} shows an example of a prompt for a parent-child local structure.

\textbf{Iterative Feedback.}
The framework elicits functional mapping in parent-child local structure without generating samples. To enforce global consistency as a DAG, it includes an iterative feedback refinement mechanism. This process requires pre-specified hard constraints for each variable, analytically performs sanity check and retry for each elicitation. Details are available in Appendix~\ref{app:framework_details} Algorithm~\ref{alg:iterative_feedback}.

\subsection{Metrics for Evaluating Parameterization}

Our framework compares ground-truth effect parameters and the effects elicited via LLMs, obtaining metrics (M1)--(M4), which we defined as in Appendix~\ref{app:metrics}. For (M1)--(M3), we compute the L2 norm over all linear coefficients across nodes; the contribution aggregation differs by metric. These metrics capture distances between LLM-elicited parameter vectors per node 
and ground truth vectors per node.
(M4) captures relative effect-size ordering per node.

\section{Experimental Setup}

We assume linear functional elicitation for pre-specified DAGs using LLMs given their nodes and directed edges.
We obtained ground-truths in linear settings, including learned parameters as well as DAG structures~\citep{LEONELLI2025129502bnrep} in the real-world settings.
In our functional elicitation, we obtain a textual structural equation directly but do not create conditional distributions from LLMs as in~\citep{bynum2025languagemodelscausaleffect}. For DAGs and their parameterization, we focus on continuous variables. The bayesian networks we used for ground-truths have learned  effects in continuous linear Gaussian settings. A small subset of DAGs, for example, some variables in \emph{Expenditure} DAG, for example, \emph{Card}, whether card is accepted or not, was also binary but because the ground-truth network learned this effect as continuous, in our prompt we also treat it as continuous. 

\subsection{LLM Model Selection}
We employed general purpose pre-trained LLMs (Gemini 2.5 Flash, Llama 3 family, and GPT-5.4) as in Appendix~\ref{app:exp_details} Table~\ref{tab:model_benchmarks}. We tried models with different model sizes and architectures either mixture-of-experts (MoE), where an input is routed into different experts, obtaining output in an ensemble fashion~\citep{jacobs1991adaptive}, vs non-MoE ones.  Gemini 2.5 Flash consists of sparse MoE~\citep{comanici2025gemini25pushingfrontier}.  Representative models from the open-weights community, Llama 3.1 8B and Llama 3.3 70B have dense transformer, where all parameters are used for inference~\citep{llama32024}. From frontier model family, we include GPT-5.4~\footnote{\url{https://openai.com/index/introducing-gpt-5-4/}} with limited model details disclosed.

\subsection{DAGs With Ground Truth Effect Parameters}

Our experiment employed real-world DAGs from a Bayesian Network repository, \emph{BnRep}~\citep{LEONELLI2025129502bnrep}. We kept the ones with a continuous Linear Gaussian setting which corresponds to the linear SCM setting.
For convenience, we included DAGs with at most 15 nodes.
We excluded DAGs where variable names consisted of letters and numbered suffixes. Appendix~\ref{app:dag_selection} Figure~\ref{fig:dag_inclusion_flowchart} describes how we selected DAGs and effect ground-truths from the repository, and Table~\ref{tab:dag_table} lists the 7 DAGs included in our experiments. %

\subsection{Adversarial Conditions}

As in RQ2, we introduce two adversarial conditions. %

\textbf{(I) Tweaking Units to Check Robustness.} For the first type of adversarial conditions, namely changes in units, we use \emph{Cachexia1} DAG from Table~\ref{tab:dag_table} because the DAG comes with units and the DAG structure is relatively simpler among others. %
We captured aggregated trends on each model with temperature set to zero to make the behavior as deterministic as possible.
We employed the following two conditions: (A) the \emph{Cachexia} DAG with original units $\mu$M, and (B) the \emph{Cachexia} DAG with tweaked units, nM.

\textbf{(II) Simulated DAG Misspecification.}

We created adversarial mutated examples based on Expenditure DAG (details in Appendix Figure~\ref{fig:expenditure_dag}). For each, we added a spurious edge between two variables with no actual connections in the ground truths while ensuring acyclicity. The resulting four examples are referred to as (S1)-(S4) as discussed in Appendix~\ref{app:dag_miss_samplers}.
Based on this, we performed parameterization to check how robust each LLM is.

\section{Results and Discussion}

In our experiments, a \emph{run} consists of a DAG traversal from the first node to the last. For each condition (DAG, LLM), we run 25 independent runs (i.e., $n=25$) and report mean M1--M4 values. Values after $\pm$ denote 95\% CIs.

\subsection{RQ1: Direct Estimation Results}
Table~\ref{tab:direct_estimation} presents direct parameter estimation results across four LLM models on all DAGs.
M1 shows high variability across DAGs because it is not scale-invariant with respect to variable ranges. We therefore focus on M2, M3, and M4. Under M2, the best model varies by DAG. Under M3, Gemini 2.5 Flash and GPT-5.4 each perform best on three DAGs. Under M4, GPT-5.4 performs best on four DAGs and Gemini 2.5 Flash on three, including ties. Comparing Llama 3.1 8B and Llama 3.3 70B, the larger model usually performs better, suggesting a benefit from larger parameter counts for this task. A small model, Llama 3.1 8B failed to generate parsable equations for \emph{algal2}.

While we used temperature zero, larger models (Gemini 2.5 Flash, Llama 3.3 70B, and GPT-5.4) still show substantial stochasticity across metrics, with non-Gaussian distributions at $n=25$, possibly due to hardware or software factors beyond our control via external APIs. This variability is concerning for safety-critical domains such as healthcare. This result aligns with prior work on LLM non-determinism~\citep{klishevich2025measuringdeterminismlargelanguage}---a failure mode in response to RQ3.

\begin{table}[h]
\caption{Direct estimation results in the main text (Averaged $n=25$, Temp 0) with 95\% CIs. We focus discussion on $M3$ and $M4$; the full $M1$--$M4$ table is provided in Appendix~\ref{app:main_res_details} Table~\ref{tab:direct_estimation_full_app}. For $M3$, lower is better ($\downarrow$); for $M4$, higher is better ($\uparrow$). See Table~\ref{tab:dag_table} for DAG descriptions.}
\label{tab:direct_estimation}
\begin{center}
\footnotesize
\setlength{\tabcolsep}{3pt} 
\begin{tabular}{llcc}
\toprule
	\textbf{MODEL} & \textbf{DAG} & \textbf{M3} $\downarrow$ & \textbf{M4} $\uparrow$ \\
\midrule
Gemini 2.5 Flash & \multirow{4}{*}{cachexia1}    & $1.07 \pm 0.11$ & $\textbf{1.00} \pm 0.00$ \\
Llama 3.1 8B     &                                 & $1.04 \pm 0.00$ & $0.00 \pm 0.00$ \\
Llama 3.3 70B    &                                 & $1.35 \pm 0.07$ & $0.44 \pm 0.20$ \\
GPT-5.4          &                                 & $\textbf{0.98} \pm 0.05$ & $0.64 \pm 0.19$ \\
\midrule
Gemini 2.5 Flash & \multirow{4}{*}{expenditure}  & $\textbf{1.00} \pm 0.23$ & $\textbf{7.52} \pm 0.20$ \\
Llama 3.1 8B     &                                 & $2.05 \pm 0.00$ & $7.00 \pm 0.00$ \\
Llama 3.3 70B    &                                 & $1.55 \pm 0.23$ & $6.56 \pm 0.20$ \\
GPT-5.4          &                                 & $1.96 \pm 0.08$ & $6.36 \pm 0.19$ \\
\midrule
Gemini 2.5 Flash & \multirow{4}{*}{foodsecurity} & $0.24 \pm 0.05$ & $0.50 \pm 0.21$ \\
Llama 3.1 8B     &                                 & $0.45 \pm 0.00$ & $0.00 \pm 0.00$ \\
Llama 3.3 70B    &                                 & $0.40 \pm 0.00$ & $0.00 \pm 0.00$ \\
GPT-5.4          &                                 & $\textbf{0.11} \pm 0.03$ & $\textbf{0.96} \pm 0.08$ \\
\midrule
Gemini 2.5 Flash & \multirow{4}{*}{algal2}       & $\textbf{0.51} \pm 0.10$ & $\textbf{2.00} \pm 0.00$ \\
Llama 3.1 8B     &                                 & \multicolumn{2}{c}{n/a} \\
Llama 3.3 70B    &                                 & $0.56 \pm 0.05$ & $\textbf{2.00} \pm 0.00$ \\
GPT-5.4          &                                 & $0.52 \pm 0.04$ & $\textbf{2.00} \pm 0.00$ \\
\midrule
Gemini 2.5 Flash & \multirow{4}{*}{lexical}      & $2.04 \pm 0.05$ & $2.04 \pm 0.27$ \\
Llama 3.1 8B     &                                 & $2.56 \pm 0.00$ & $1.00 \pm 0.00$ \\
Llama 3.3 70B    &                                 & $2.27 \pm 0.04$ & $0.78 \pm 0.37$ \\
GPT-5.4          &                                 & $\textbf{1.76} \pm 0.07$ & $\textbf{2.24} \pm 0.20$ \\
\midrule
Gemini 2.5 Flash & \multirow{4}{*}{liquefaction} & $\textbf{0.84} \pm 0.03$ & $2.80 \pm 0.16$ \\
Llama 3.1 8B     &                                 & $2.01 \pm 0.00$ & $\textbf{3.00} \pm 0.00$ \\
Llama 3.3 70B    &                                 & $1.27 \pm 0.03$ & $\textbf{3.00} \pm 0.00$ \\
GPT-5.4          &                                 & $0.88 \pm 0.02$ & $2.96 \pm 0.08$ \\
\midrule
Gemini 2.5 Flash & \multirow{4}{*}{stocks}       & $1.01 \pm 0.06$ & $3.48 \pm 0.24$ \\
Llama 3.1 8B     &                                 & $1.59 \pm 0.00$ & $2.88 \pm 0.13$ \\
Llama 3.3 70B    &                                 & $\textbf{0.93} \pm 0.06$ & $2.60 \pm 0.23$ \\
GPT-5.4          &                                 & $1.18 \pm 0.06$ & $\textbf{3.80} \pm 0.12$ \\
\bottomrule
\end{tabular}
\end{center}
\end{table}

\subsection{RQ2: Robustness Results}

\textbf{(I) Unit Tweaking Robustness.}
Appendix~\ref{app:main_res_details} Table~\ref{tab:unit_tweak_robustness_results} evaluates model robustness when unit values for the Cachexia1 DAG are tweaked from $\mu$M to nM. Here we only provide overview.
In the lens of M2-M4, sometimes, counter-intuitively tweaked units resulted in better parameterization. One possible reason that the metric improved instead of degradation is increased numerical precision in textual formal, e.g., \emph{1} $\mu$M $\rightarrow$ \emph{1000} nM. We plan to extend the empirical coverage in the future to see if this is the case. 
\todo[inline]{[ICML WS TODO]: 1 $\mu$M $\rightarrow$ 1000 nM, then internal reasoning steps might have more precision, resulting in better performance? But to test this, I need another example with milli, 1 mM $\rightarrow$ 1000 $\mu$M, so the precision could decrease]}

\todo[inline]{[ICML WS TODO]: Why tweaked results are generally better? Did I include beta 0 intercept in metric calculation? If so, for intercept, when I tweak things, I should also scale intercept?}
\todo[inline]{[ICML WS TODO]: will add two more unit purtubatioon dags.}

\textbf{(II) DAG Misspecification Robustness.} Table~\ref{tab:dag_misspecification_robustness_results} presents the performance when the Expenditure DAG is intentionally misspecified by adding spurious edges, where O indicates the original DAG while S1-S4 indicate mutated variants; the full $M1$--$M4$ table is provided in Appendix~\ref{app:main_res_details} Table~\ref{tab:dag_misspecification_robustness_full_app}. 
As above, we focus on M3 and M4 in the main text. Under M3, the original DAG yields the best result for Gemini 2.5 Flash, Llama 3.1 8B, and Llama 3.3 70B, while GPT-5.4 performs best on S1. Most models therefore degrade under DAG misspecification by M3. Under M4, the original DAG performs best for Gemini 2.5 Flash and Llama 3.1 8B. For Llama 3.3 70B and GPT-5.4, variant S4 
is slightly better than the original DAG, which remains second best.
\todo[inline]{[GR TODO]: S4 is an exeption. From structure, S4 is also unique, Selfemp had only one parent in original DAG but in S4, another link is added to Selfemp, making Selfemp with multiple parents. Can that contribute to this exceptin in M4?}
\todo[inline]{[ICML WS TODO]: We need to calculate supression metric instead of M4 for more detailed discussion}
Overall, adversarial conditions tend to lower M4, indicating degraded performance under DAG misspecification with spurious edges---a failure mode regarding RQ3.

\begin{table}[t]
\caption{Robustness under DAG misspecification in the main text for the expenditure DAG (Averaged $n=25$, Temp 0) with 95\% CIs. We focus discussion on $M3$ and $M4$; the full $M1$--$M4$ table is provided in Appendix~\ref{app:main_res_details} Table~\ref{tab:dag_misspecification_robustness_full_app}. For $M3$, lower is better ($\downarrow$); for $M4$, higher is better ($\uparrow$). Misspec: (O)=Original DAG, (S1)-(S4)=Spurious edges added.}
\label{tab:dag_misspecification_robustness_results}
\begin{center}
\footnotesize
\setlength{\tabcolsep}{3pt}
\begin{tabular}{lccc}
\toprule
	\textbf{MODEL} & \textbf{MISSPEC.} & \textbf{M3} $\downarrow$ & \textbf{M4} $\uparrow$ \\
\midrule
Gemini 2.5 Flash & O & $\textbf{1.00} \pm 0.23$ & $\textbf{7.52} \pm 0.20$ \\
                 & S1 & $1.37 \pm 0.28$ & $6.20 \pm 0.30$ \\
                 & S2 & $1.24 \pm 0.27$ & $6.40 \pm 0.25$ \\
                 & S3 & $1.35 \pm 0.25$ & $6.56 \pm 0.26$ \\
                 & S4 & $1.99 \pm 0.19$ & $7.32 \pm 0.27$ \\
\midrule
Llama 3.1 8B     & O & $\textbf{2.05} \pm 0.00$ & $\textbf{7.00} \pm 0.00$ \\
                 & S1 & $2.48 \pm 0.00$ & $6.00 \pm 0.00$ \\
                 & S2 & $2.06 \pm 0.00$ & $6.00 \pm 0.00$ \\
                 & S3 & $2.11 \pm 0.00$ & $6.00 \pm 0.00$ \\
                 & S4 & $2.41 \pm 0.00$ & $\textbf{7.00} \pm 0.00$ \\
\midrule
Llama 3.3 70B    & O & $\textbf{1.55} \pm 0.23$ & $6.56 \pm 0.20$ \\
                 & S1 & $1.70 \pm 0.22$ & $5.52 \pm 0.23$ \\
                 & S2 & $1.92 \pm 0.19$ & $5.92 \pm 0.25$ \\
                 & S3 & $2.04 \pm 0.12$ & $5.36 \pm 0.19$ \\
                 & S4 & $2.22 \pm 0.14$ & $\textbf{6.68} \pm 0.27$ \\

\midrule
GPT-5.4            & O & $1.96 \pm 0.08$ & $6.36 \pm 0.19$ \\
                   & S1 & $\textbf{0.90} \pm 0.03$ & $6.00 \pm 0.00$ \\
                   & S2 & $1.02 \pm 0.08$ & $6.00 \pm 0.00$ \\
                   & S3 & $1.01 \pm 0.10$ & $6.12 \pm 0.13$ \\
                   & S4 & $1.74 \pm 0.09$ & $\textbf{6.96} \pm 0.08$ \\
\bottomrule
\end{tabular}
\end{center}
\end{table}

\section{Conclusion}

Linear-LLM-SCM benchmarks quantitative causal effect elicitation by decomposing DAGs into local structures, along with four evaluation metrics. Across four LLMs, Gemini 2.5 Flash performs best overall on M3 and M4 across the tested DAGs. These results highlight the importance of scale-invariant metrics such as M4, since M1 is affected by variable ranges. A challenge is the stochasticity of three large LLMs, which produce inconsistent results at temperature zero. In contrast, a small LLM was more consistent but less accurate. Such variance poses risks in safety-critical domains, e.g., healthcare. Robustness tests show that adding spurious edges degrades performance and lowers effect-ordering accuracy, indicating limited robustness to structural uncertainty. Future work should extend it to non-linear functional forms and investigate ways to mitigate structural noise. We did not utilize confidence intervals from LLMs, but plan to examine whether such textual uncertainty outputs are  calibrated.

\section*{Impact Statement}

We foresee our work being useful to benchmark LLMs for causal effect elicitation in safety-critical domains, such as healthcare, where quantitative causal expert knowledge is needed. There are several downstream societal implications of using LLMs but we do not specify all such implications in this manuscript. Given the current results from the manuscript, we need to clarify that we should not use LLMs for quantitative causal effect estimation for real healthcare or clinical decision support, but still we need further research on safety testing and improvements. 

\section*{Acknowledgement}

We acknowledge funding for the project AI4Nof1 by the state of Rhineland Palatinate, Germany. We would like to thank Valentin Margraf, Jonas Hanselle, Serafima Lebedeva and Niklas Nertinger for their valuable feedback during weekly research meeting.

\todo[inline]{TODO: add this paragraph (required)}

\bibliography{example_paper}
\bibliographystyle{icml2026}

\newpage
\appendix
\onecolumn

\section*{Appendix}

\section{Background and Related Work}
\label{app:relatedwork}

The section first describes the theoretical foundation
of structural causal models.   
Then, we will examine general LLMs' abilities and inabilities reported in the causality, which is a broad area of study including causal discovery and causal effect estimation. Then we will focus on existing efforts of using LLMs for causal effect estimation in the causality literature. 

\subsection{Structural Causal Models and Effect Parameterization}

A Structural Causal Model is formally defined by a pair $(\mathcal{G}, \mathbf{F})$, where $\mathcal{G}$ is the Directed Acyclic Graph (DAG) representing the causal structure, and $\mathbf{F} = \{f_i\}$ is the collection of structural equations. Each endogenous variable $X_i$ is determined by a function $f_i$ of its direct causes (parents) $\text{Pa}(X_i)$ and an independent exogenous noise term $E_i$, such that $X_i = f_i(\text{Pa}(X_i), E_i)$. In our problem settings, we focus on linear causal effects to narrow down the problem space, and therefore, parameterization here refers to finding coefficients of these linear functions $\mathbf{F}$. 

Conventional causal inference tasks include finding causal relationships, direction of edges and identifying such parameters using data and intervention in the real-world. However, in our work, we only aim to estimate such parameters for linear causal effects using potentially encoded knowledge from large language models (LLMs) for each variable in a \emph{pre-specified} DAG for a real-world phenomenon.

\subsubsection{LLMs General Potential and Criticism for Causality}

For causal discovery, \cite{long2024llmsbuildcausal, long2023causaldiscoverylanguagemodels} used LLMs to identify causal connections between node pairs in DAGs, reporting opportunities despite inconsistencies and prompt sensitivities. \cite{kıcıman2024causalreasoninglargelanguage} found LLM-based methods outperform covariance-based algorithms in pairwise causal discovery and excel at natural language counterfactual reasoning.

There is also work raising limitations of the current LLMs in this regard. For example, \citet{zhang2023understandingcausalitylargelanguage} proposed LLMs' Causal Hierarchy, which consists of the three types: \emph{Type 1: Identifying causal relationships using domain knowledge}, \emph{Type 2: Discovering new knowledge from data}, \emph{Type 3: Quantitative estimation of consequences of actions}. They claim LLMs can perform Type 1 tasks via training data but not Type 2 and Type 3 due to token generation limitations. Our work evaluates LLMs' capability for Type 3 tasks. Another study by \citet{zecevic2023causalparrotslargelanguage} raises doubts about LLMs' causal capabilities, conjecturing they merely learned causal facts from training data rather than causal mechanisms, \emph{Causal Parrots}. They note LLMs are not explicitly trained for causal tasks and may simply parrot causal statements without true reasoning. Their criticism focuses on causal discovery; since we assume given causal structures and only estimate effects, this may not directly apply to our settings. 
Criticisms also include ground truth leakage in causal discovery~\cite{yang2023criticalrevcausalinfbenchllm}. In the broader context of causal inference, \citet{jin2024cladderassessingcausalreasoning} introduced CausalCOT, a prompting strategy for the whole causal reasoning lifecycle, concluding this task is highly challenging for LLMs. 

\subsubsection{LLMs for Causal Effect Estimation}

While some literature indicates initial success reports for treatment effect estimation in \textbf{discrete} domains, including counterfactual generation, causal effect estimation in \textbf{continuous} domains (e.g., SCMs) remains largely unexplored. Most existing approaches for continuous causal estimation provide numerical observational data to the model, which often requires specialized architectures or tokenization schemes; by contrast, we study coefficient elicitation without observational data, using only DAG structure and variable semantics.

\citet{bynum2025languagemodelscausaleffect} combined LLMs with structural causal modeling for pre-specified DAGs. However, they use sampling-based approaches to estimate effects rather than directly eliciting functional mappings. Their SD-SCMs represent effects as Conditional Probability Distributions rather than linear structural equations~\citep{Wright1921CorrelationAndCausation}, focus on discrete-domain counterfactuals, while our work addresses continuous-domain parametrization for simulating real-world phenomena. \citet{liu2025llmandcausalinferencecolab} surveys LLM-based causal inference, finding only three papers on treatment effect estimation—two in discrete domains~\citep{chen2023distilincounterfactualswithllms, feder2024dataaugmentationsimprovedlarge} and one~\citep{zhang2024causalfoundationmodelduality} addressing both domains but requiring numerical observational data with specialized attention mechanisms. Unlike these approaches, we assume pre-specified causal DAG structures without observational data. \citet{nafar2025extractingprobabilisticknowledgelarge} similarly benchmark \emph{effect} estimation from pre-specified DAGs but assume discrete domains with Conditional Probability and sampling distributions, using eighty public DAGs in finance and health. Other benchmarks~\citep{wang2024causalbench, mohan2025llmsunderstanddrugmechanisms} examine LLMs on causality tasks but focus on causal relationship identification rather than quantitative effect estimation in continuous SCMs. %

\section{Benchmarking Framework Implementation Details}
\label{app:benchmark_impl_details}

\subsection{Plug-and-play Configuration Example}
\label{app:dag_config_yaml}
\begin{figure}[h]
\begin{center}
\includegraphics[width=\columnwidth]{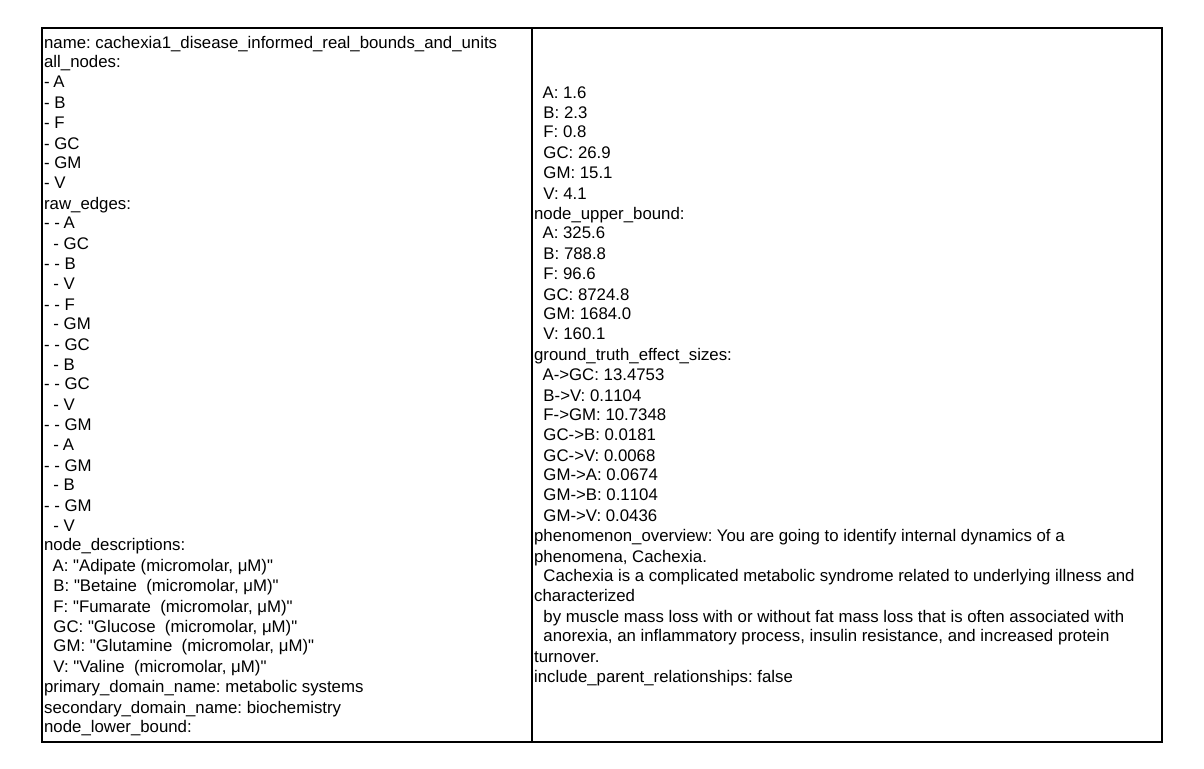}
\end{center}
\caption{An example configuration of plug-and-play DAG yaml for our open-source framework.}    \label{fig:yaml_dag_configuration}
\end{figure}

Our open-source tooling requires a very simple yaml format to specify DAG structures and other inputs required as shown in Figure~\ref{fig:yaml_dag_configuration}.

\subsection{Example Output Response from an LLM to a Prompt}
\label{app:income_prompt_and_output}
\begin{figure}[h]
\begin{center}
\includegraphics[width=\columnwidth]{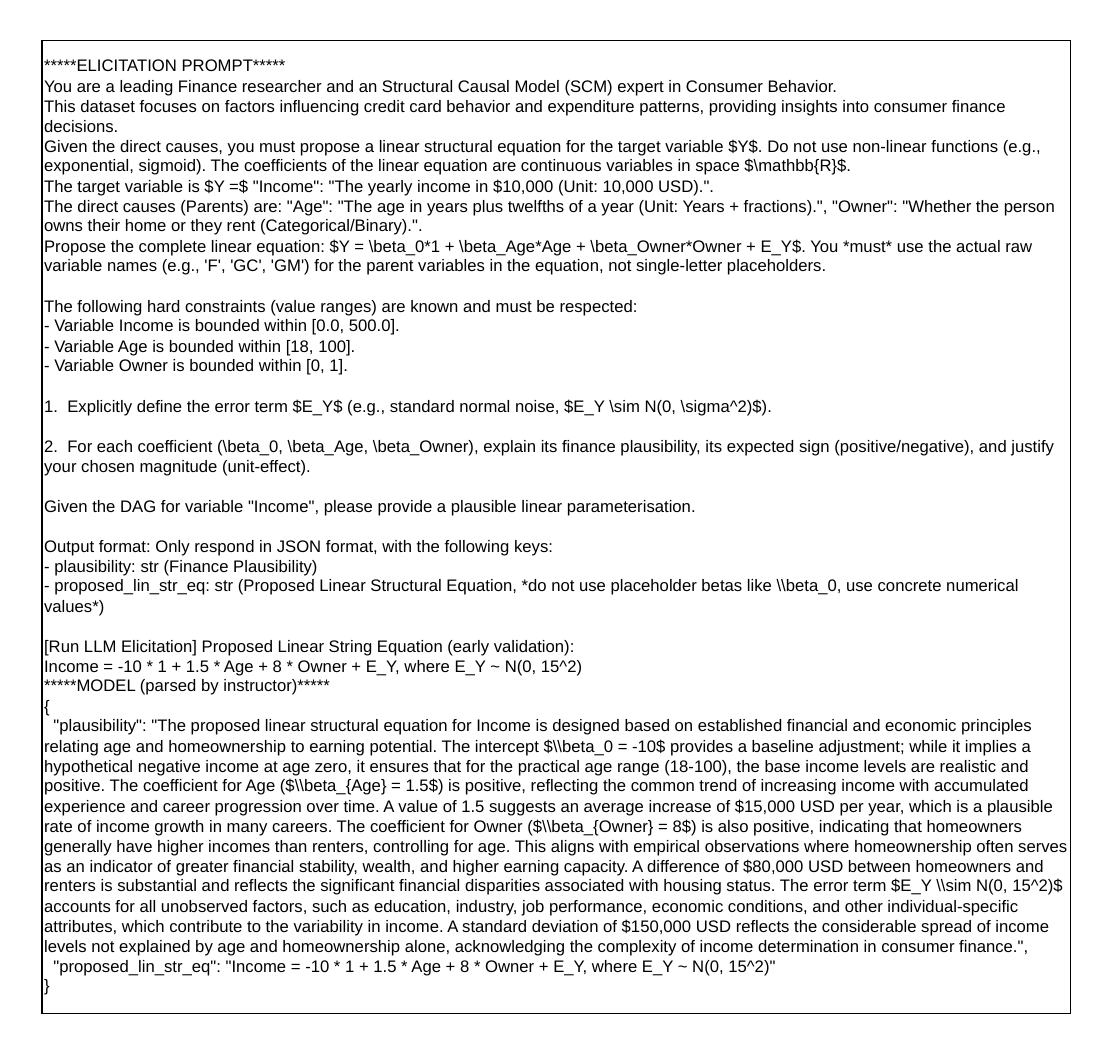}
\end{center}
\caption{An example prompt and its LLM response structure at the Expenditure DAG at \emph{Income} target variable. The output can be found at the bottom as JSON structured format.}    \label{fig:income_prompt_and_output}
\end{figure}

Please refer to Figure~\ref{fig:income_prompt_and_output}. 

\section{Benchmarking Framework Details}
\label{app:framework_details}

\subsection{Algorithm Details}
\label{app:algo_details}

Algorithm~\ref{alg:benchmarking} provides details of the main parts of the framework.

\begin{algorithm}[tb]
   \caption{Linear-LLM-SCM Benchmarking Framework}
   \label{alg:benchmarking}
\begin{algorithmic}[1]
   \REQUIRE DAG structure $\mathcal{G}=(\mathcal{V}, \mathcal{E})$, Variable descriptions $\mathcal{D}$, Value ranges $\mathcal{R}$, Phenomenon overview $\mathcal{P}$, LLM $\mathcal{M}$
   \ENSURE Aggregated set of linear coefficients $\boldsymbol{\hat{\beta}}$
   
   \STATE Initialize $\boldsymbol{\hat{\beta}} \leftarrow \emptyset$
   \FOR{each target variable $X_j \in \mathcal{V}$ in topological order}
      \STATE 1. Identify direct parents $Pa(X_j) \subset \mathcal{V}$ using edges $\mathcal{E}$ (decompose into local structures)
      \STATE 2. Construct prompt $S_j$ incorporating:
      \STATE \quad \textbullet \enspace Domain expert persona and Phenomenon overview $\mathcal{P}$
      \STATE \quad \textbullet \enspace Short descriptions $\mathcal{D}$ and Units for $X_j$ and $Pa(X_j)$
      \STATE \quad \textbullet \enspace Hard constraints/Value ranges $\mathcal{R}$ 
      \STATE 3. Call LLM: $Response \leftarrow \mathcal{M}(S_j)$ requesting JSON format
      \STATE 4. Parse $Response$: Extract numerical $\hat{\beta}$ for $Pa(X_j)$ and intercept $\beta_0$ (with iterative feedback, see Algorithm~\ref{alg:iterative_feedback})
      \STATE 5. $\boldsymbol{\hat{\beta}} \leftarrow \boldsymbol{\hat{\beta}} \cup \{\hat{\beta}_{i,j}\}$
   \ENDFOR
   \STATE \textbf{return} aggregated parameterized SCM $\boldsymbol{\hat{\beta}}$ for graph structure $\mathcal{G}$
\end{algorithmic}
\end{algorithm}

\subsection{Prompt Example for Parent-Child Elicitation}
\label{app:prompt_template}

\begin{figure}[h]
\begin{center}
\includegraphics[width=\columnwidth]{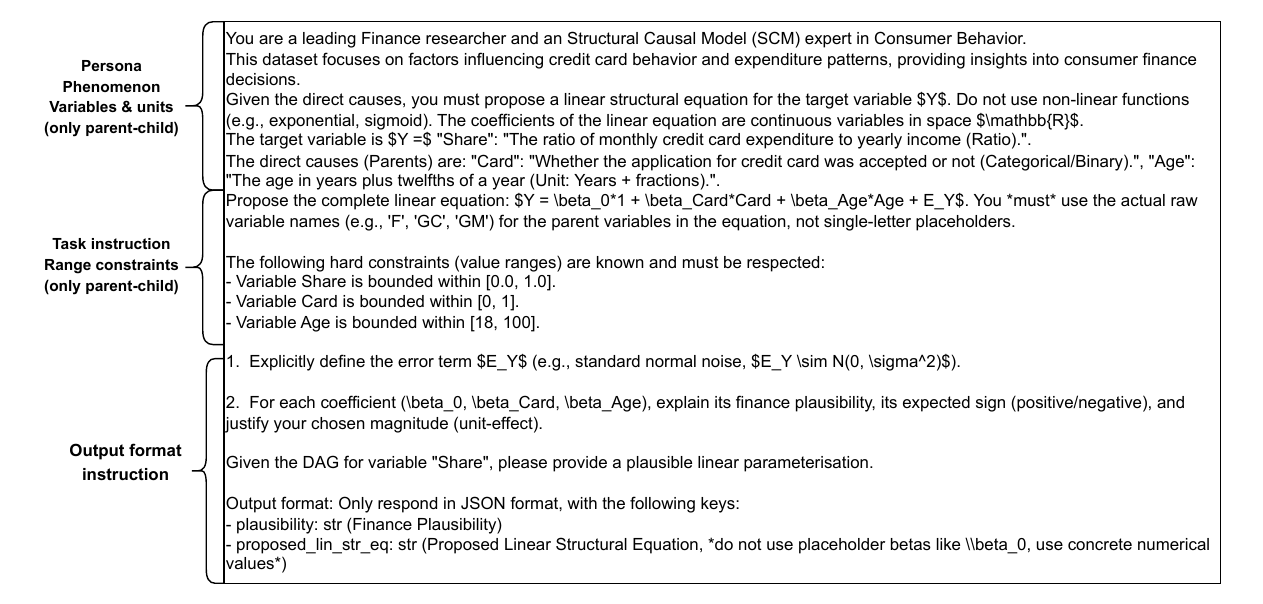} 
\end{center}
\caption{An example of a prompt for a local parent-child structure in a DAG. }
\label{fig:prompt_template}
\end{figure}

An example is shown in Figure~\ref{fig:prompt_template}.

\subsection{The iterative hard-constraint loop overview}
\label{app:iterative_details}

\begin{algorithm}[ht]
\caption{Iterative Feedback for Refinement with Hard Constraints}
\label{alg:iterative_feedback}
\begin{algorithmic}[1]
\REQUIRE Target variable $X_j$, Parent hard constraints $R_{Pa(X_j)}$, Node hard constraints $C_2$, Loop budget $n$
\ENSURE Accepted linear coefficients $\hat{\beta}$

\FOR{$\textit{iteration} = 1$ \textbf{to} $n$}
    \STATE Call LLM $\mathcal{M}$ with prompt $S_j$ to get proposal $P$ (includes parameterization $\hat{\beta}$)
    \STATE Calculate possible value range $C_1$ of $X_j$ based on $R_{Pa(X_j)}$ and $P$ 
    
    \IF{$C_2 \text{ includes } C_1$} 
        \STATE \textbf{return} $\hat{\beta}$ (accept proposal) 
    \ELSE
        \STATE Reject proposal 
        \STATE Update prompt $S_j$ with previous proposal and validation results 
    \ENDIF
\ENDFOR
\STATE \textbf{return} last available $\hat{\beta}$ (budget $n$ reached) 
\end{algorithmic}
\end{algorithm}

\todo[inline]{TODO: explain the iterative feedback mechanism here}

\begin{figure}[h]
\begin{center}
\includegraphics[width=\columnwidth]{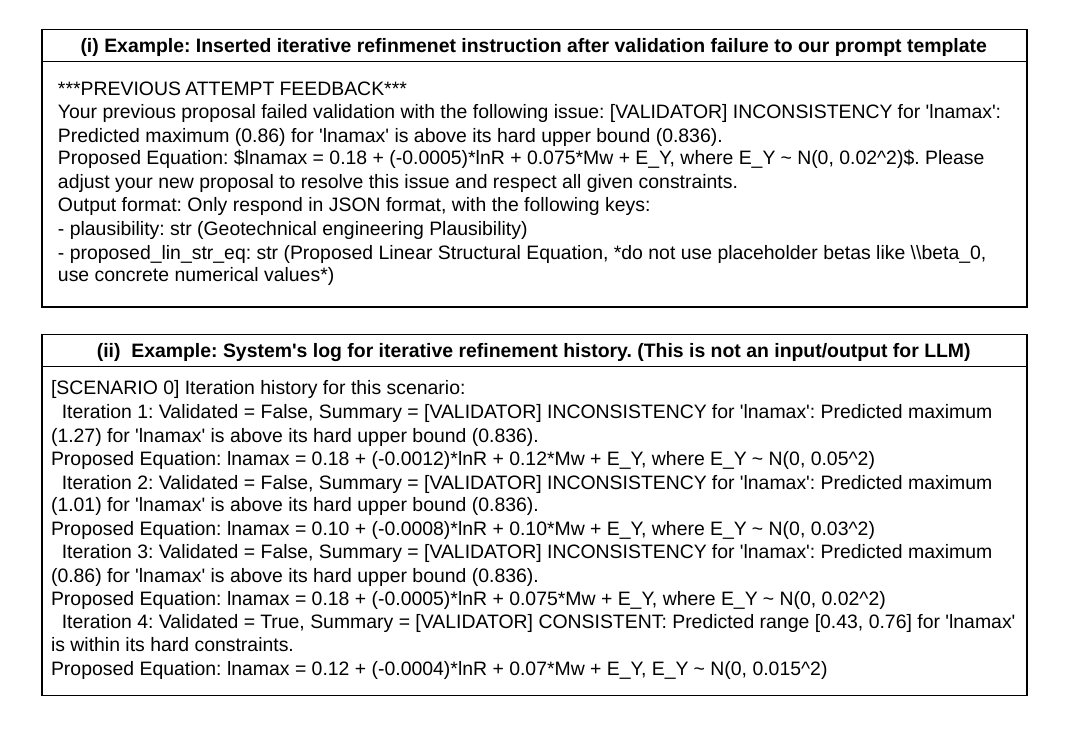}
\end{center}
\caption{(i) Example of refinement instruction inserted to the next iteration's prompt and (ii) iterative refinement history over time from the system's log.}    \label{fig:iterative_template}
\end{figure}

In addition to the Algorithm~\ref{alg:iterative_feedback}, here we provide Figure~\ref{fig:iterative_template} to show a concrete example of (i) how feedback on failure is incooperated to the next iteration's prompt, and (ii) an example of iterative refinement history from our system's log. As in (i) example, the previously proposed but failed proposal and its failure reason is inserted at the last position of the standard prompt template in the next iteration. Please note that, if the proposal fails multiple times, only the last proposal was provided to the next iteration.  The example (ii) demonstrates an example, where a variable \emph{lnamax} violated constraint in the first proposal, iteratively refineing the proposal at the end after this process. 

\todo[inline]{Maybe the way we calculate the taget variable interval based on parent intervals and parent effect estimates? TODO: write down here or integarate to algo 2 table.also Do we really need iterative feedback? I don't know if I have logs for iterative feedback count really. at least json file has if number 5 is depleted or not, something I can aggregate but let's avoid additional expeirmtn for this}

\subsection{The iterative hard-constraint loop details}

In our parameterization, the LLM aims to obtain the following structural equation for each node. In practice, while our system prompts up to the noise $\varepsilon$, but we do not utilize or examine the values, which is future work. 

\begin{equation}
  Y = \beta_0 + \sum_{i=1}^{n} \beta_i X_i + \varepsilon
  \label{eq:linear_scm}
\end{equation}

Given the elicited coefficients $\beta_0, \beta_1, \dots, \beta_n$ and the known bounds
$[lb_i,\, ub_i]$ for each parent variable $X_i$, the validator computes the worst-case
deterministic range $[\hat{Y}_{\min},\, \hat{Y}_{\max}]$ of $Y$ using interval arithmetic. 

\subsubsection{Initialization}

Both extremes are initialised with the intercept:
\begin{equation}
  \hat{Y}_{\min} = \beta_0, \qquad \hat{Y}_{\max} = \beta_0
\end{equation}

\subsubsection{Parent Contribution}

For each parent variable $X_i$ with coefficient $\beta_i$ and bounds $[lb_i,\, ub_i]$:

\begin{equation}
  \hat{Y}_{\min} \mathrel{+}=
  \begin{cases}
    \beta_i \cdot lb_i & \text{if } \beta_i \geq 0 \\
    \beta_i \cdot ub_i & \text{if } \beta_i < 0
  \end{cases}
  \qquad
  \hat{Y}_{\max} \mathrel{+}=
  \begin{cases}
    \beta_i \cdot ub_i & \text{if } \beta_i \geq 0 \\
    \beta_i \cdot lb_i & \text{if } \beta_i < 0
  \end{cases}
  \label{eq:interval_arith}
\end{equation}

\subsubsection{Validation Check}

Let $[lb_Y,\, ub_Y]$ be the hard bounds on the target variable $Y$.
The proposed equation \emph{passes} validation if and only if:
\begin{equation}
  lb_Y \leq \hat{Y}_{\min} \quad \text{and} \quad \hat{Y}_{\max} \leq ub_Y
  \label{eq:validation_condition}
\end{equation}

If either condition is violated the validator returns a human-readable failure message, e.g.\
\textit{``Predicted minimum ($-0.30$) for `$Y$' is below its hard lower bound ($0.01$).''}

The noise term $\varepsilon$ is excluded from the interval
calculation because at this point, we do not know how reliably LLMs can perform uncertainty quantification in a symbolic fashion, which is future work. 

\section{Mathematical Definitions for Our Metrics}
\label{app:metrics}

These metrics capture distances between LLM-elicited parameter vectors per node $\boldsymbol{\beta}_{LLM,j}$ and ground truth vectors per node $\boldsymbol{\beta}_{GT,j}$. For each vector, we denote $j$'s each direct parents' edges effect size (scalar) using index $i$: $\beta_{LLM,j,i}$ and $\beta_{GT,j,i}$. The metrics are summarized as follows:

\begin{itemize}
    \item \textbf{(M1) L2 Norm distance} between LLM-elicited vs GT:
    \begin{equation}\label{eq:1}
        M_{1}=\sqrt{\sum_{j \in V} \sum_{i \in Pa(j)} (\beta_{LLM, j, i} - \beta_{GT, j, i})^2}
    \end{equation}
    \item \textbf{(M2) L2 Norm distance} with node-wise effect normalization:
    \begin{equation}\label{eq:2}
        M_{2}=\sqrt{\sum_{j \in V} \sum_{i \in Pa(j)} \left( \frac{\beta_{LLM, j, i}}{\|\boldsymbol{\beta}_{LLM, j}\|_2} - \frac{\beta_{GT, j, i}}{\|\boldsymbol{\beta}_{GT, j}\|_2} \right)^2}
    \end{equation}
    \item \textbf{(M3) L2 Norm distance} excluding edges with single parents:
    \begin{equation}\label{eq:3}
        M_3 = \sqrt{\sum_{j \in \{V : |Pa(X_j)| > 1\}} \sum_{i \in Pa(X_j)} \left( \frac{\beta_{LLM, j, i}}{\|\boldsymbol{\beta}_{LLM, j}\|_2} - \frac{\beta_{GT, j, i}}{\|\boldsymbol{\beta}_{GT, j}\|_2} \right)^2}
    \end{equation}
    \item \textbf{(M4) Effect size relative ordering} matches per target variable:
    \begin{equation}\label{eq:4}
        M_{4}=\sum_{j\in\{V:|Pa(X_{j})|>1\}}\mathbb{I}\left(\text{ordering of } \{\beta_{\text{LLM}, i, j}\}_{i \in Pa(X_j)} = \text{ordering of } \{\beta_{\text{GT}, i, j}\}_{i \in Pa(X_j)}\right)
    \end{equation}
\end{itemize}

(M4) captures relative effect-size ordering per node. If a node $j$ has parents $a$ and $b$ and the effect sizes are ${\beta}_{LLM,j,a}=-0.8 < {\beta}_{LLM,j,b}=0.5$ while  ${\beta}_{GT,j,a}=-2 < {\beta}_{GT,j,b}=3$, for this node we increment the sum by one before moving to the next node, finally obtaining (M4). 

\section{Experimental Details}
\label{app:exp_details}

\subsection{LLM Model Details}

Please refer to Table~\ref{tab:model_benchmarks}.

\begin{table}[t]
\caption{Benchmark model specifications for our study. MoE stands for mixture of experts. Dense means dense-transformer and this also indicates the model is non-MoE-based. The tick \checkmark indicates yes, and the cross $\times$ indicates no.}
\label{tab:model_benchmarks}
\begin{center}
\begin{tabular}{llll}
\toprule
\multicolumn{1}{c}{\bf MODEL NAME}  &\multicolumn{1}{c}{\bf MODEL SIZE} &\multicolumn{1}{c}{\bf ARCHITECTURE} &\multicolumn{1}{c}{\bf OPEN WEIGHTS}
\\ \midrule
Gemini 2.5 Flash & Unknown & MoE   & $\times$ \\
Llama 3.1 8B     & 8B & Dense & \checkmark \\
Llama 3.3 70B & 70B & Dense & \checkmark \\ %
GPT-5.4 & Unknown & Unknown & $\times$ %

\\ \bottomrule
\end{tabular}
\end{center}
\end{table}

\subsection{LLM Interaction Protocol}

While the full experimental pipeline is publicly available and open-source in
~\footnote{[URL reducted for double-blind review]}
, here we describe key details for reproducibility.  

In all models, we set temperature to zero while we did not change top k / top p sampling because generally when setting temperature, it is recommended to fall back to default top k and top p values. For Groq inference API, there is an option to explicitly set seed, but we do not explicitly set this value. For Gemini API, there is also a similar entry to specify the seed but we did not explicitly set this value. For both providers, setting this value does not guarantee actual seed being set to the specified value because they describe this as best-effort~\footnote{\url{https://console.groq.com/docs/api-reference\#chat-create}}~\footnote{\url{https://docs.cloud.google.com/vertex-ai/generative-ai/docs/model-reference/inference}}, and non-deterministic also seem issues acknowledged by the model providers. For OpenAI GPT-5.4 models, we set \texttt{reasoning\_effort} to \texttt{none} to enable setting temperature to zero.

For Gemini models, via Google AI Studio's APIs. For Llama models via Groq APIs. All the APIs are called via a python package called instructor~\footnote{\url{https://python.useinstructor.com/concepts/validation/\#3-custom-validators}}. Rate limiting was handled using the instructor package with its default exponential backoff. The JSON format was also enforced via the instructor package. 

\todo[inline]{[ICML WS TODO]: Run small case study with fixed seed and see if high variance can be mitigated. If so, our claim about variance needs to change. The variance decomposition is also mentioned by the reviewers.}

\section{DAG Descriptions}
\label{app:dag_selection}

\subsection{DAGs used for experiment}

Please refer to Table~\ref{tab:dag_table}.

\subsection{DAGs selection flowchart}

\begin{figure}[h]
\begin{center}
\includegraphics[width=0.6\columnwidth]{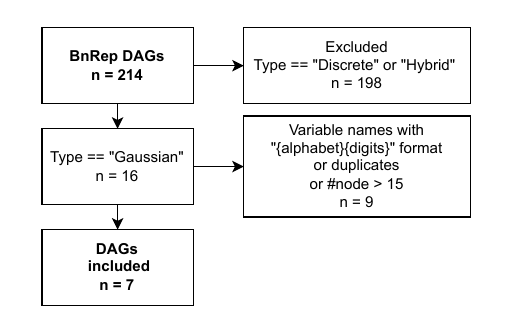}
\end{center}
\caption{Inclusion and exclusion flowchart for DAG ground-truths from BnRep DAG repository}    \label{fig:dag_inclusion_flowchart}
\end{figure}

Please refer to Figure~\ref{fig:dag_inclusion_flowchart}.

\begin{table}[b] 
\caption{DAGs with ground-truths included for our study from BnRep repository. \textbf{Literature} indicates sources introducing either the DAG structure or DAG effect parameterization. \textbf{VR.} (Value Ranges) indicates whether value ranges are available in the original literature. }
\label{tab:dag_table}
\begin{center}
\begin{tabular}{llllll}
\toprule
\multicolumn{1}{c}{\bf NAME} & \multicolumn{1}{c}{\bf LITERATURE} & \multicolumn{1}{c}{\bf VR.} & \multicolumn{1}{c}{\bf NODES} & \multicolumn{1}{c}{\bf DOMAIN}
\\ \midrule
cachexia1    & \citet{gorgen_model-preserving_2020, eisner_learning_2011}  & Y & 6  & Genetics \\ %
expenditure  & \citet{tsagris_fedhc_2022, greene2003econometric} & N & 12 & Economics \\ %
foodsecurity & \citet{leonelli_coherent_2020, barons_eliciting_2018} &Y & 4  & Social Sciences \\ %
algal2       &  \citet{jackson-blake_seasonal_2022} & Y & 9  & Env. Science \\ %
lexical      & \citet{baumann_accounting_2022} & Y & 8  & Social Sciences \\ %
liquefaction & \citet{hu_continuous_2023} & Y & 10 & Earth Sciences \\ %
stocks       & \citet{sener_gaussian_2024} & Y & 13 & Economics \\ %

\bottomrule
\end{tabular}
\end{center}
\end{table}

\subsection{Samples of DAGs Used in the Experiments}

\begin{figure}[ht]
    \centering
    \includegraphics[width=\textwidth]{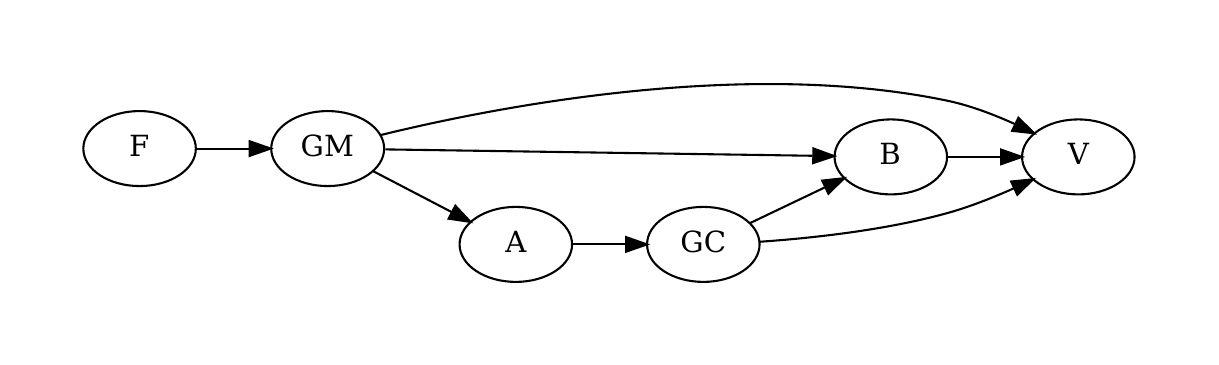}
    \caption{The DAG structure of cachexia1 from BnRep repository.}
    \label{fig:cachexia_dag}
\end{figure}

\begin{figure}[ht]
    \centering
    \includegraphics[width=\textwidth]{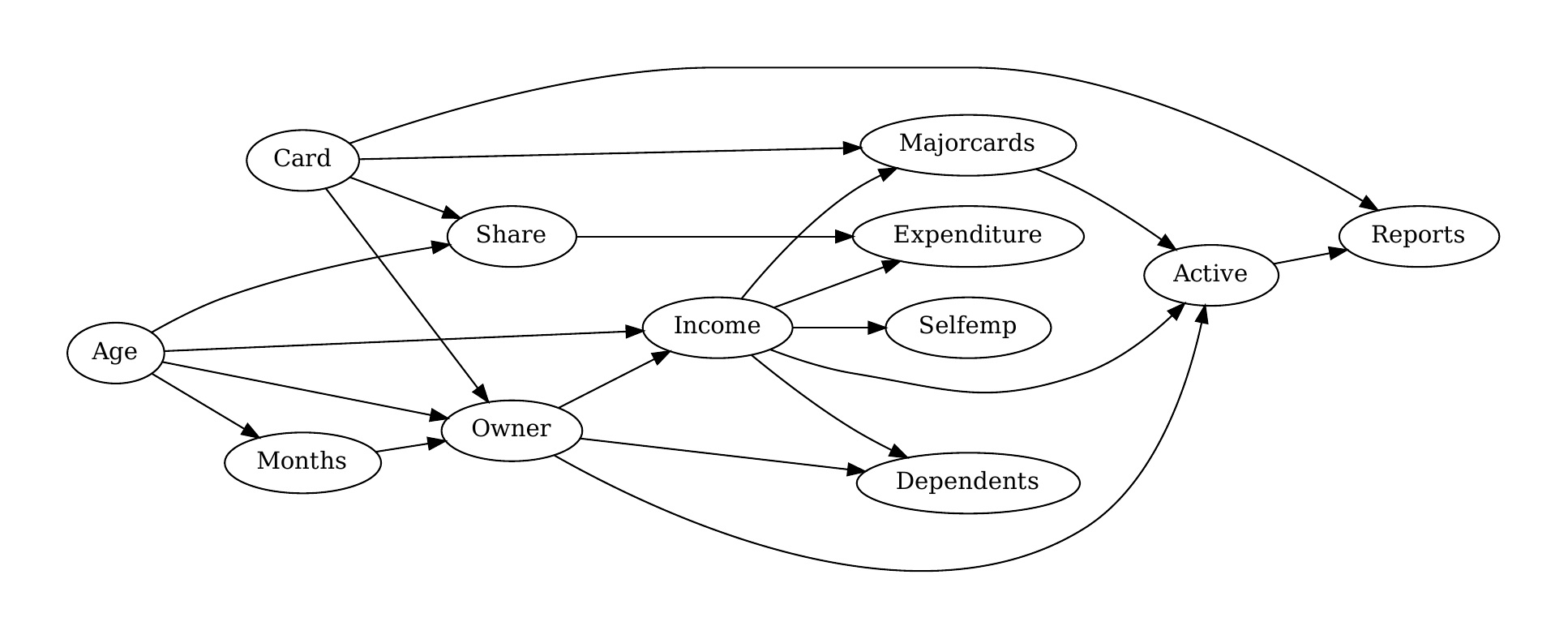}
    \caption{The DAG structure of expenditure from BnRep repository.}
    \label{fig:expenditure_dag}
\end{figure}

One may easily find DAGs we used in our experiments from references, but to save time of the readers, we attached the two example DAGs, Figure~\ref{fig:cachexia_dag} and Figure~\ref{fig:expenditure_dag}.

\subsection{DAG misspecification samplers}
\label{app:dag_miss_samplers}
For the DAG misspecification experiment in the main text, the four mutated variants of the original Expenditure DAG add the following spurious edges: (S1) \emph{Owner} $\rightarrow$ \emph{Expenditure}, (S2) \emph{Majorcards} $\rightarrow$ \emph{Dependents}, (S3) \emph{Owner} $\rightarrow$ \emph{Share}, and (S4) \emph{Majorcards} $\rightarrow$ \emph{Selfemp}.

\section{Main Experimental Results Details}
\label{app:main_res_details}

Table~\ref{tab:direct_estimation_full_app} reports the full direct-estimation results ($M1$--$M4$) corresponding to the simplified main-text Table~\ref{tab:direct_estimation}.

\begin{table*}[t]
\caption{Full direct estimation results (Averaged $n=25$, Temp 0) with 95\% CIs. $M1$: L2 distance; $M2$: normalized L2; $M3$: normalized L2 excluding single-parent edges; $M4$: Effect relative order count. For $M1$--$M3$, lower is better ($\downarrow$); for $M4$, higher is better ($\uparrow$). See Table~\ref{tab:dag_table} for DAG descriptions.}
\label{tab:direct_estimation_full_app}
\begin{center}
\footnotesize
\setlength{\tabcolsep}{3pt} 
\begin{tabular}{llcccc}
\toprule
	MODEL & DAG & M1 $\downarrow$ & M2 $\downarrow$ & M3 $\downarrow$ & M4 $\uparrow$ \\ 
\midrule
Gemini 2.5 Flash & cachexia1 & $\textbf{13.783} \pm 1.826$ & $2.449 \pm 0.240$ & $1.074 \pm 0.114$ & $\textbf{1.000} \pm 0.000$ \\
Llama 3.1 8B     & cachexia1 & $16.527 \pm 0.000$ & $\textbf{1.036} \pm 0.000$ & $1.036 \pm 0.000$ & $0.000 \pm 0.000$ \\
Llama 3.3 70B    & cachexia1 & $13.843 \pm 0.938$ & $1.994 \pm 0.213$ & $1.352 \pm 0.071$ & $0.440 \pm 0.199$ \\
GPT-5.4          & cachexia1 & $17.043 \pm 0.069$ & $1.479 \pm 0.295$ & $\textbf{0.976} \pm 0.053$ & $0.640 \pm 0.192$ \\
\addlinespace
Gemini 2.5 Flash & expenditure & $148084.559 \pm 22359.281$ & $\textbf{0.998} \pm 0.226$ & $\textbf{0.998} \pm 0.226$ & $\textbf{7.520} \pm 0.200$ \\
Llama 3.1 8B     & expenditure & $\textbf{2463.377} \pm 0.000$ & $2.053 \pm 0.000$ & $2.053 \pm 0.000$ & $7.000 \pm 0.000$ \\
Llama 3.3 70B    & expenditure & $27137.540 \pm 12574.354$ & $1.548 \pm 0.234$ & $1.548 \pm 0.234$ & $6.560 \pm 0.199$ \\
GPT-5.4          & expenditure & $83376.520 \pm 26482.403$ & $1.960 \pm 0.076$ & $1.960 \pm 0.076$ & $6.360 \pm 0.192$ \\
\addlinespace
Gemini 2.5 Flash & foodsecurity & $22.801 \pm 0.038$ & $2.017 \pm 0.006$ & $0.236 \pm 0.050$ & $0.500 \pm 0.214$ \\
Llama 3.1 8B     & foodsecurity & $22.727 \pm 0.006$ & $\textbf{0.447} \pm 0.000$ & $0.447 \pm 0.000$ & $0.000 \pm 0.000$ \\
Llama 3.3 70B    & foodsecurity & $22.914 \pm 0.009$ & $2.039 \pm 0.001$ & $0.395 \pm 0.004$ & $0.000 \pm 0.000$ \\
GPT-5.4          & foodsecurity & $\textbf{22.587} \pm 0.044$ & $1.248 \pm 0.371$ & $\textbf{0.110} \pm 0.028$ & $\textbf{0.960} \pm 0.078$ \\
\addlinespace
Gemini 2.5 Flash & algal2 & $\textbf{4.094} \pm 0.256$ & $\textbf{0.514} \pm 0.098$ & $\textbf{0.514} \pm 0.098$ & $\textbf{2.000} \pm 0.000$ \\
Llama 3.1 8B     & algal2 & \multicolumn{4}{c}{\emph{Model output equations not parsable by the program.}} \\
Llama 3.3 70B    & algal2 & $4.657 \pm 0.004$ & $0.559 \pm 0.054$ & $0.559 \pm 0.054$ & $\textbf{2.000} \pm 0.000$ \\
GPT-5.4          & algal2 & $4.376 \pm 0.058$ & $0.523 \pm 0.039$ & $0.523 \pm 0.039$ & $\textbf{2.000} \pm 0.000$ \\
\addlinespace
Gemini 2.5 Flash & lexical & $42.246 \pm 0.650$ & $2.036 \pm 0.053$ & $2.036 \pm 0.053$ & $2.040 \pm 0.265$ \\
Llama 3.1 8B     & lexical & $35.854 \pm 0.000$ & $3.252 \pm 0.000$ & $2.565 \pm 0.000$ & $1.000 \pm 0.000$ \\
Llama 3.3 70B    & lexical & $35.847 \pm 0.001$ & $2.274 \pm 0.037$ & $2.274 \pm 0.037$ & $0.778 \pm 0.374$ \\
GPT-5.4          & lexical & $\textbf{35.122} \pm 2.235$ & $\textbf{1.759} \pm 0.072$ & $\textbf{1.759} \pm 0.072$ & $\textbf{2.240} \pm 0.205$ \\
\addlinespace
Gemini 2.5 Flash & liquefaction & $12.294 \pm 5.556$ & $\textbf{0.844} \pm 0.034$ & $\textbf{0.844} \pm 0.034$ & $2.800 \pm 0.160$ \\
Llama 3.1 8B     & liquefaction & $9.999 \pm 0.000$ & $2.012 \pm 0.002$ & $2.012 \pm 0.002$ & $\textbf{3.000} \pm 0.000$ \\
Llama 3.3 70B    & liquefaction & $12.093 \pm 0.221$ & $1.271 \pm 0.026$ & $1.271 \pm 0.026$ & $\textbf{3.000} \pm 0.000$ \\
GPT-5.4          & liquefaction & $\textbf{4.400} \pm 0.453$ & $0.881 \pm 0.022$ & $0.881 \pm 0.022$ & $2.960 \pm 0.078$ \\
\addlinespace
Gemini 2.5 Flash & stocks & $0.829 \pm 0.048$ & $1.059 \pm 0.122$ & $1.006 \pm 0.063$ & $3.478 \pm 0.242$ \\
Llama 3.1 8B     & stocks & $0.893 \pm 0.005$ & $1.589 \pm 0.003$ & $1.589 \pm 0.003$ & $2.880 \pm 0.130$ \\
Llama 3.3 70B    & stocks & $1.060 \pm 0.026$ & $\textbf{0.934} \pm 0.059$ & $\textbf{0.934} \pm 0.059$ & $2.600 \pm 0.226$ \\
GPT-5.4          & stocks & $\textbf{0.566} \pm 0.014$ & $0.936 \pm 0.021$ & $0.936 \pm 0.021$ & $\textbf{3.680} \pm 0.187$ \\
\bottomrule
\end{tabular}
\end{center}
\end{table*}

Table~\ref{tab:unit_tweak_robustness_results} reports the full unit-tweaking robustness results referenced in the main text.

\begin{table}[t]
\caption{Robustness under unit tweak (Averaged $n=25$, Temp 0) with 95\% CIs. $M1$: L2 distance; $M2$: normalized L2; $M3$: normalized L2 excluding single-parent edges; $M4$: Effect relative order count. For $M1$--$M3$, lower is better ($\downarrow$); for $M4$, higher is better ($\uparrow$). Units: (L)=$\mu$M, (T)=nM.}
\label{tab:unit_tweak_robustness_results}
\begin{center}
\footnotesize
\setlength{\tabcolsep}{3pt} 
\begin{tabular}{llcccccc}
\toprule
	\textbf{MODEL} & \textbf{CND.} & \textbf{UNITS} & \textbf{M1} $\downarrow$ & \textbf{M2} $\downarrow$ & \textbf{M3} $\downarrow$ & \textbf{M4} $\uparrow$ \\ 
\midrule
Gemini 2.5 Flash & A & L & $13.783 \pm 1.826$ & $2.449 \pm 0.240$ & $1.074 \pm 0.114$ & $1.000 \pm 0.000$ \\
                 & B & T & $\textbf{12.399} \pm 1.774$ & $\textbf{1.769} \pm 0.372$ & $\textbf{0.917} \pm 0.136$ & $\textbf{1.120} \pm 0.206$ \\
\addlinespace
Llama 3.1 8B     & A & L & $\textbf{16.527} \pm 0.000$ & $1.036 \pm 0.000$ & $1.036 \pm 0.000$ & $\textbf{0.000} \pm 0.000$ \\
                 & B & T & $17.198 \pm 0.000$ & $\textbf{0.905} \pm 0.002$ & $\textbf{0.905} \pm 0.002$ & $\textbf{0.000} \pm 0.000$ \\
\addlinespace
Llama 3.3 70B    & A & L & $13.843 \pm 0.938$ & $\textbf{1.994} \pm 0.213$ & $1.352 \pm 0.071$ & $0.440 \pm 0.199$ \\
                 & B & T & $\textbf{13.441} \pm 0.199$ & $2.439 \pm 0.160$ & $\textbf{1.107} \pm 0.115$ & $\textbf{0.640} \pm 0.250$ \\
\addlinespace
GPT-5.4 & A & L & $17.043 \pm 0.069$ & $1.479 \pm 0.295$ & $0.976 \pm 0.053$ & $0.640 \pm 0.192$ \\
                 & B & T & $\textbf{8.086} \pm 0.763$ & $\textbf{0.952} \pm 0.129$ & $\textbf{0.903} \pm 0.073$ & $\textbf{1.000} \pm 0.000$ \\
\bottomrule
\end{tabular}
\end{center}
\end{table}

Table~\ref{tab:dag_misspecification_robustness_full_app} reports the full DAG-misspecification robustness results ($M1$--$M4$) corresponding to the simplified main-text Table~\ref{tab:dag_misspecification_robustness_results}.

\begin{table*}[t]
\caption{Full robustness under DAG misspecification for the expenditure DAG (Averaged $n=25$, Temp 0) with 95\% CIs. $M1$: L2 distance; $M2$: normalized L2; $M3$: normalized L2 excluding single-parent edges; $M4$: Effect relative order count. For $M1$--$M3$, lower is better ($\downarrow$); for $M4$, higher is better ($\uparrow$). Misspec: (O)=Original DAG, (S1)-(S4)=Spurious edges added.}
\label{tab:dag_misspecification_robustness_full_app}
\begin{center}
\footnotesize
\setlength{\tabcolsep}{3pt}
\begin{tabular}{lcccccc}
\toprule
	\textbf{MODEL} & \textbf{MISSPEC.} & \textbf{M1} $\downarrow$ & \textbf{M2} $\downarrow$ & \textbf{M3} $\downarrow$ & \textbf{M4} $\uparrow$ \\
\midrule
Gemini 2.5 Flash & O & $148084.559 \pm 22359.281$ & $\textbf{0.998} \pm 0.226$ & $\textbf{0.998} \pm 0.226$ & $\textbf{7.520} \pm 0.200$ \\
                 & S1 & $\textbf{81587.682} \pm 20626.822$ & $1.372 \pm 0.282$ & $1.372 \pm 0.282$ & $6.200 \pm 0.299$ \\
                 & S2 & $138770.541 \pm 21002.249$ & $1.235 \pm 0.267$ & $1.235 \pm 0.267$ & $6.400 \pm 0.253$ \\
                 & S3 & $120240.669 \pm 20548.802$ & $1.355 \pm 0.253$ & $1.355 \pm 0.253$ & $6.560 \pm 0.255$ \\
                 & S4 & $106012.716 \pm 23004.372$ & $1.988 \pm 0.189$ & $1.988 \pm 0.189$ & $7.320 \pm 0.271$ \\
\addlinespace
Llama 3.1 8B     & O & $\textbf{2463.377} \pm 0.000$ & $\textbf{2.053} \pm 0.000$ & $\textbf{2.053} \pm 0.000$ & $\textbf{7.000} \pm 0.000$ \\
                 & S1 & $20151.134 \pm 0.000$ & $2.483 \pm 0.000$ & $2.483 \pm 0.000$ & $6.000 \pm 0.000$ \\
                 & S2 & $\textbf{2463.377} \pm 0.000$ & $2.063 \pm 0.001$ & $2.063 \pm 0.001$ & $6.000 \pm 0.000$ \\
                 & S3 & $\textbf{2463.377} \pm 0.000$ & $2.106 \pm 0.000$ & $2.106 \pm 0.000$ & $6.000 \pm 0.000$ \\
                 & S4 & $\textbf{2463.377} \pm 0.000$ & $2.413 \pm 0.000$ & $2.413 \pm 0.000$ & $\textbf{7.000} \pm 0.000$ \\
\addlinespace
Llama 3.3 70B    & O & $27137.540 \pm 12574.354$ & $\textbf{1.548} \pm 0.234$ & $\textbf{1.548} \pm 0.234$ & $6.560 \pm 0.199$ \\
                 & S1 & $25843.000 \pm 13086.141$ & $1.703 \pm 0.217$ & $1.703 \pm 0.217$ & $5.520 \pm 0.230$ \\
                 & S2 & $29202.638 \pm 13299.788$ & $1.918 \pm 0.186$ & $1.918 \pm 0.186$ & $5.920 \pm 0.251$ \\
                 & S3 & $24257.613 \pm 11979.590$ & $2.038 \pm 0.118$ & $2.038 \pm 0.118$ & $5.360 \pm 0.192$ \\
                 & S4 & $\textbf{12524.332} \pm 7698.439$ & $2.222 \pm 0.138$ & $2.222 \pm 0.138$ & $\textbf{6.680} \pm 0.271$ \\
\addlinespace
GPT-5.4          & O & $\textbf{83376.520} \pm 26482.403$ & $1.960 \pm 0.076$ & $1.960 \pm 0.076$ & $6.360 \pm 0.192$ \\
                 & S1 & $132030.159 \pm 25271.455$ & $\textbf{0.902} \pm 0.025$ & $\textbf{0.902} \pm 0.025$ & $6.000 \pm 0.000$ \\
                 & S2 & $107136.488 \pm 28206.749$ & $1.020 \pm 0.077$ & $1.020 \pm 0.077$ & $6.000 \pm 0.000$ \\
                 & S3 & $91970.954 \pm 28957.569$ & $1.010 \pm 0.095$ & $1.010 \pm 0.095$ & $6.120 \pm 0.130$ \\
                 & S4 & $100496.485 \pm 26558.895$ & $1.738 \pm 0.085$ & $1.738 \pm 0.085$ & $\textbf{6.960} \pm 0.078$ \\
\bottomrule
\end{tabular}
\end{center}
\end{table*}

\section{Prompt Sensitivity Analysis for Structural Awareness}

In our framework, prompt has the highest degree of freedom. While it is costly to try all the ideas, because LLMs are very sensitive prompts, for example, even new lines or separators affects the next token prediction~\cite{sclar2024quantifying}, it makes sense to revisit our prompts. For our additional experiments, due to budget constraints, we only performed experiments with three models, Gemini 2.5 Flash, Llama 3.1 8B and Llama 3.3 70B. 

\textbf{Local structure awareness in prompt:}

\begin{figure}[h]
\begin{center}
\includegraphics[width=\columnwidth]{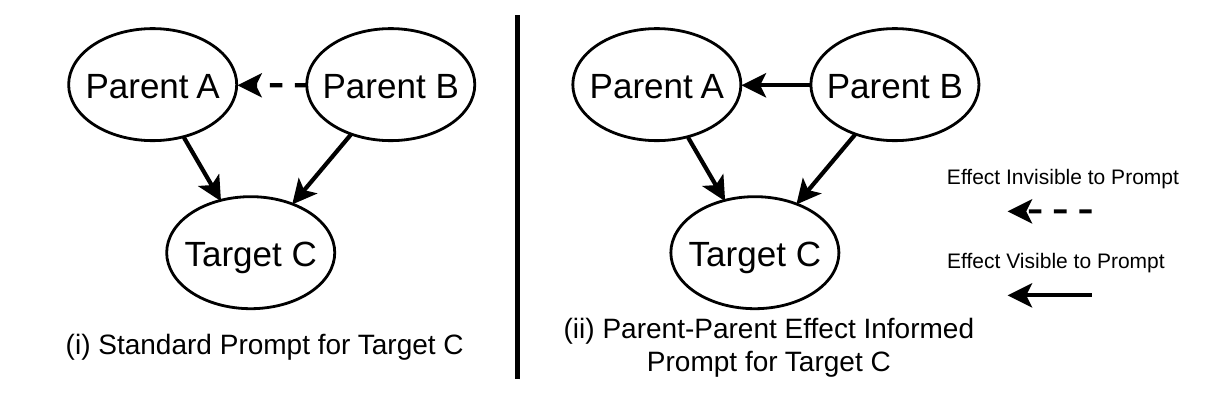}
\end{center}
\caption{Comparison of (i) standard prompt and (ii) parent-parent effect informed with the simplest structural example where parents relate to each other.}    \label{fig:parent_parent_prompt}
\end{figure}

The default prompt we introduced is a local parent-child set with only visibility on parent-child effect relationship, which is miopic structure awareness. We also devised another variant of prompt, ``parent-parent-effect informed'', which includes parent-parent relationship in addition to the existing effect information in prompts. Figure~\ref{fig:parent_parent_prompt} visualizes a simple example of (i) the standard prompt and (ii) parent-parent effect informed prompt. Regardless of this additional information, in the Linear SCM formulation, the direct effect should be the same in theory. With this prompt with slightly less miopic example, we would like to see the reaction of LLMs in terms of direct effect estimation. 

We summarized the result for parent-parent-effect informed prompt in Table~\ref{tab:direct_estimation_parent_parent_prompt}. Overall, compared to the standard prompt, in most cases, parent-parent-effect informed prompt yielded worse performances, justifying the standard prompt we used in our main experiments. 

\begin{table}[t]
\caption{Direct estimation results with parent-parent effect prompting (Averaged $n=25$, Temp 0). Values after $\pm$ indicate 95\% CIs. $M1$: L2 distance; $M2$: normalized L2; $M3$: normalized L2 excluding single-parent edges; $M4$: Effect relative order count. For $M1$--$M3$, lower is better ($\downarrow$); for $M4$, higher is better ($\uparrow$). See Table~\ref{tab:dag_table} for DAG descriptions.}
\label{tab:direct_estimation_parent_parent_prompt}
\begin{center}
\footnotesize
\setlength{\tabcolsep}{3pt} 
\begin{tabular}{llcccc}
\toprule
	\textbf{MODEL} & \textbf{DAG} & \textbf{M1} $\downarrow$ & \textbf{M2} $\downarrow$ & \textbf{M3} $\downarrow$ & \textbf{M4} $\uparrow$ \\ 
\midrule
Gemini 2.5 Flash     & cachexia        & $\mathbf{13.100} \pm 2.247$ & $2.059 \pm 0.339$ & $\mathbf{0.803} \pm 0.224$ & $\mathbf{1.440} \pm 0.199$ \\
Llama 3.1 8B         & cachexia        & $16.527 \pm 0.000$ & $\mathbf{1.047} \pm 0.000$ & $1.047 \pm 0.000$ & $0.000 \pm 0.000$ \\
Llama 3.3 70B        & cachexia        & $14.424 \pm 0.786$ & $1.774 \pm 0.262$ & $1.200 \pm 0.092$ & $0.400 \pm 0.196$ \\
\addlinespace
Gemini 2.5 Flash     & expenditure     & $125421.804 \pm 24860.693$ & $1.729 \pm 0.216$ & $1.729 \pm 0.216$ & $6.560 \pm 0.279$ \\
Llama 3.1 8B         & expenditure     & $\mathbf{2463.302} \pm 0.000$ & $2.099 \pm 0.001$ & $2.099 \pm 0.001$ & $6.000 \pm 0.000$ \\
Llama 3.3 70B        & expenditure     & $19137.678 \pm 10242.559$ & $\mathbf{1.711} \pm 0.208$ & $\mathbf{1.711} \pm 0.208$ & $\mathbf{6.680} \pm 0.246$ \\
\addlinespace
Gemini 2.5 Flash     & foodsecurity    & $22.852 \pm 0.035$ & $2.049 \pm 0.011$ & $\mathbf{0.422} \pm 0.058$ & $\mathbf{0.080} \pm 0.108$ \\
Llama 3.1 8B         & foodsecurity    & $\mathbf{22.700} \pm 0.029$ & $\mathbf{0.628} \pm 0.013$ & $0.628 \pm 0.013$ & $0.000 \pm 0.000$ \\
Llama 3.3 70B        & foodsecurity    & $22.982 \pm 0.025$ & $2.049 \pm 0.000$ & $0.447 \pm 0.000$ & $0.000 \pm 0.000$ \\
\addlinespace
Gemini 2.5 Flash     & algal2          & $\mathbf{4.203} \pm 0.307$ & $\mathbf{0.572} \pm 0.172$ & $\mathbf{0.572} \pm 0.172$ & $\mathbf{2.000} \pm 0.000$ \\
Llama 3.1 8B         & algal2          & $4.961 \pm 0.000$ & $1.079 \pm 0.001$ & $1.079 \pm 0.001$ & $\mathbf{2.000} \pm 0.000$ \\
Llama 3.3 70B        & algal2          & $4.635 \pm 0.039$ & $0.614 \pm 0.036$ & $0.614 \pm 0.036$ & $\mathbf{2.000} \pm 0.000$ \\
\addlinespace
Gemini 2.5 Flash     & lexical         & $41.733 \pm 2.350$ & $\mathbf{1.963} \pm 0.061$ & $\mathbf{1.963} \pm 0.061$ & $\mathbf{2.480} \pm 0.230$ \\
Llama 3.1 8B         & lexical         & $\mathbf{1.609} \pm 0.000$ & $2.923 \pm 0.000$ & $2.131 \pm 0.000$ & $2.000 \pm 0.000$ \\
Llama 3.3 70B        & lexical         & $10.442 \pm 0.023$ & $2.232 \pm 0.053$ & $2.232 \pm 0.053$ & $2.040 \pm 0.211$ \\
\addlinespace
Gemini 2.5 Flash     & liquefaction    & $13.087 \pm 4.536$ & $\mathbf{0.850} \pm 0.028$ & $\mathbf{0.850} \pm 0.028$ & $2.840 \pm 0.147$ \\
Llama 3.1 8B         & liquefaction    & $\mathbf{9.998} \pm 0.000$ & $0.985 \pm 0.002$ & $0.985 \pm 0.002$ & $\mathbf{3.000} \pm 0.000$ \\
Llama 3.3 70B        & liquefaction    & $11.745 \pm 0.079$ & $1.302 \pm 0.036$ & $1.302 \pm 0.036$ & $\mathbf{3.000} \pm 0.000$ \\
\addlinespace
Gemini 2.5 Flash     & stocks          & $\mathbf{0.786} \pm 0.043$ & $1.194 \pm 0.111$ & $1.145 \pm 0.070$ & $3.040 \pm 0.309$ \\
Llama 3.1 8B         & stocks          & $0.831 \pm 0.000$ & $0.945 \pm 0.001$ & $0.945 \pm 0.001$ & $\mathbf{3.960} \pm 0.078$ \\
Llama 3.3 70B        & stocks          & $1.060 \pm 0.023$ & $\mathbf{0.935} \pm 0.037$ & $\mathbf{0.935} \pm 0.037$ & $2.680 \pm 0.187$ \\
\addlinespace

\bottomrule
\end{tabular}
\end{center}
\end{table}

\todo[inline]{TODO: this is the most priority!}

\end{document}